# Optical Neural Ordinary Differential Equations

YUN ZHAO[1,3,6], HANG CHEN[2,6], MIN LIN[4], HAIOU ZHANG[2], TAO YAN[1], XING LIN[2,5,7], RUQI HUANG[3,7], AND QIONGHAI DAI[1,5,7]

[1] Department of Automation, Tsinghua University, Beijing 100084, China.

[2] Department of Electronic Engineering, Tsinghua University, Beijing 100084, China

[3] Tsinghua-Berkeley Shenzhen Institute, Tsinghua University, Shenzhen University Town, Guangdong 518071, China

[4] Sea AI Lab, 1 Fusionopolis Place, 17-10, Galaxis 138522, Singapore

[5] Beijing National Research Center for Information Science and Technology, Tsinghua University, Beijing 100084, China

[6] Co-first author

[7] Emails: lin-x@tsinghua.edu.cn, ruqihuang@sz.tsinghua.edu.cn, daiqh@tsinghua.edu.cn

## ABSTRACT

Increasing the layer number of on-chip photonic neural networks (PNNs) is essential to improve its model performance. However, the successively cascading of network hidden layers results in larger integrated photonic chip areas. To address this issue, we propose the optical neural ordinary differential equations (ON-ODE) architecture that parameterizes the continuous dynamics of hidden layers with optical ODE solvers. The ON-ODE comprises the PNNs followed by the photonic integrator and optical feedback loop, which can be configured to represent residual neural networks (ResNet) and recurrent neural networks with effectively reduced chip area occupancy. For the interference-based optoelectronic nonlinear hidden layer, the numerical experiments demonstrate that the single hidden layer ON-ODE can achieve approximately the same accuracy as the two-layer optical ResNet in image classification tasks. Besides, the ON-ODE improves the model classification accuracy for the diffraction-based all-optical linear hidden layer. The time-dependent dynamics property of ON-ODE is further applied for trajectory prediction with high accuracy.

Computing processors, such as field programmable gate array (FPGA) [1], application-specific integrated circuits (ASIC) [2], and graphics or tensor processing units (GPU/TPU) [3], have been rapidly developed in recent years to implement deep neural networks for their applications in scientific research and industrial fields. With the approaching of a physical limit of electronic transistors, it's necessary to develop new computing modalities [4, 5]. Photonic computing, using photons instead of electrons for computation, has attracted increasing attention worldwide. By propagating through optical elements, photons could perform mathematical calculations at the speed of light with low energy cost, giving rise to a promising next-generation computing processor. Various PNN architectures have been proposed based on the fundamental principles of photonic spike processing [6–8], optical reservoir computing [9–12], and optical scatter materials [13]. Besides, optical diffraction [14–17] and optical interference [18–20] have been adopted to implement large-scale interconnections between neural network hidden layers.

Despite the success of on-chip PNNs in performing artificial intelligence (AI) tasks, including object classification, salient object detection [21], and vowel recognition [18], it highly relies on the hidden layer number, i.e., the depth of the network. In general, deepening neural networks was considered as a way of improving model performance. However, deploying deeper neural networks to the photonic integrated circuit requires a larger chip area. Therefore, it would be of great importance for us to optimize and improve the model performance in performing AI tasks given a limited chip area size.

It has been verified that parameterizing the continuous dynamics of neural network hidden layer with ordinary differential equations (ODE) can achieve higher accuracy in image classification under the same numbers of network hidden layers and parameters [22]. The ODE-based neural network allows sharing of the parameters among different layers. Given the target model performance, the ODE-based neural networks can be applied to reduce the models' size with higher memory efficiency effectively. Since the on-chip area size of the photonic integrated circuit is directly determined by the number of optical devices specified by neural network parameters, the ODE-based PNNs can improve the model performance and reduce the chip area size when deploying a trained model on hardware.

In this work, we propose a novel on-chip ODE-based PNN architecture, termed optical neural ordinary differential equations (ON-ODE), which comprises the photonic neural networks (PNNs) and integrators with optical feedback loop. The ON-ODE can perform object classification tasks with higher model accuracy and reduce the occupied on-chip area size. We verify the proposed approach by configuring the ON-ODE with Mach–Zehnder interferometer (MZI)-based optoelectronic nonlinear PNNs [19] and diffractive photonic computing units (DPU)-based all-optical linear PNNs respectively. The all-optical solver for ordinary differential equations [23–27] is achieved by using the ultra-fast add-drop type micro-ring resonator [28] to implement the photonic integrators and on-chip optical waveguides to implement the optical feedback loop. The numerical evaluations demonstrate the effectiveness of the proposed ON-ODE over conventional PNNs.

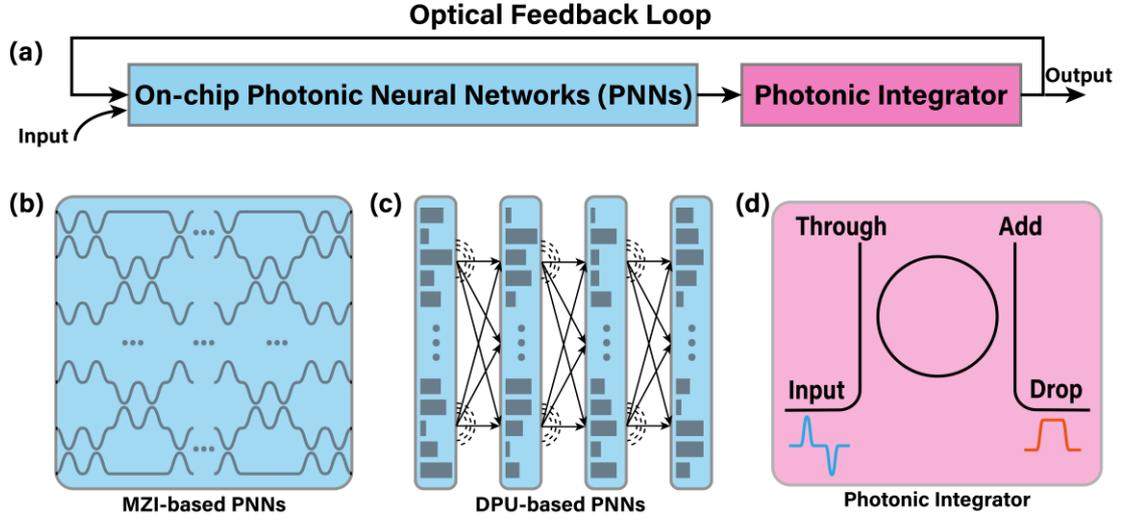

**Fig. 1.** The basic architecture of ON-ODE comprises the on-chip PNNs, photonic integrator, and optical feedback loop (a), where the PNNs are implemented with the MZI-based (b) and DPU-based (c) PNNs and the integrator is implemented with the add-drop micro-ring resonator (d).

The neural ODE consists of neural network layers with an ODE solver to parameterize the continuous-time dynamics of hidden layers [22], which can be formulated as

$$\frac{d\boldsymbol{h}(t)}{dt} = f(\boldsymbol{h}(t), t, \theta), \qquad (1)$$

where $\boldsymbol{h}(t)$ denotes the continuous-time hidden states with $t$ being the time dimension, in optical field, it denotes the optical signal in the waveguide; $f(.)$ denotes the transformation function of neural networks, which can be realized by DPU or MZI in optical structure. By integrating both sides of the equation, we have:

$$\boldsymbol{h}(t) = \int f(\boldsymbol{h}(t), t, \theta) \, dt, \qquad (2)$$

which can be optically implemented with the ON-ODE architecture depicted in Fig. 1(a). The output states of on-chip PNNs are integrated over different time steps by using photonic integrators, which is then fed back to the architecture inputs. The fundamental mechanism serves as an optical ODE solver to obtain the continuous transformation of neural network states in Eq. 1. In this work, the ON-ODE is verified by using both the MZI-based [19] and DPU-based [16] fully-connected hidden layers with optoelectronic nonlinear activation and all-optical linear configuration, respectively (see Fig. 1(b,c)). The output optical waveguides ports of PNNs are connected with on-chip micro-ring resonators and the output signals are normalized and sent back to the input ports of PNNs. The input data are pre-processed into a 1D vector and encoded into the amplitude of optical pulses. We measure the intensity of the output pulses to obtain the inference results. Different components for constructing and training ON-ODE are detailed as follows.

For the MZI-based optoelectronic PNNs, the optical fully-connected layer can be implemented by using the MZIs to formulate optical meshes (see Fig. 1(b)), where a single MZI contains two beam splitters and two phase shifters, corresponding to a 2 × 2 unitary matrix. The size of a single MZI would be 0.28 μm × 1 μm. By converting a small portion of the optical signals into voltage to be applied to MZIs, the optoelectronic nonlinear activation is achieved and can be re-programmed to synthesize a ReLU-like nonlinear response.

For the DPU-based all-optical PNNs, the on-chip DPU module comprises a stack of multi-layer metaline structures. Each metaline is a 1D etched rectangle silica slot array in the silicon membrane of silicon-on-insulator substrate that forms as diffractive meta-atoms. Different geometry structures of meta-atoms determine the phase or amplitude modulation coefficients of the on-chip optical field. We implement the on-chip DPU module in this work by successively cascading four layers of metalines. We set up one metaline with a width of 0.3 μm and set the feedback waveguide to 0.22 μm height and 0.45 μm width. Given the parameters above, we could measure that a single time step of DPU-based all optical PNNs can be less than 1 ps. As to MZI-based optoelectronic PNNs, because of the optical-to-electrical time delay, time step would be around 120 ps.

We adopt the on-chip optical add-drop type micro-ring resonator as the optical integrator to achieve the accumulation of photons and complete the integration of optical signals over time (see Fig. 1(d)). The optical signals are fed into the input ports, and the integrated signals are read from the drop ports. We define the radius of the micro-ring in the middle is 47.5 μm at a wavelength of around 1550 nm. Besides, the optical waveguide delay lines are designed to guarantee the same feedback distances from output ports to input ports. For example, in DPU based PNN, the difference of feedback distance in waveguides between the neighbouring ports would be 48 μm (port distance is 24 μm), so the optical waveguide delay lines need to be designed as 48 μm.

With the above-mentioned components and devices connected with on-chip optical waveguides, the ON-ODE are constructed and trained to perform different machine learning tasks optically. The adjoint method is used to train the continuous-time ON-ODE models with constant memory cost as a function of depth [22]. The ON-ODE can be configured to implement the optical residual neural networks or recurrent neural networks for the inference task of classification and trajectory prediction, respectively. This can be inferred from the discretized version of Eq. 2:
$$\boldsymbol{h}(T+1) = \boldsymbol{h}(T) + f(\boldsymbol{h}(T), T, \theta), \tag{3}$$
where $T \in \{0, ..., N\}$ represents different time steps. Therefore, the inference results can be obtained by measuring the steady-state response for classification tasks or the evolving system states for time-dependent predictions.

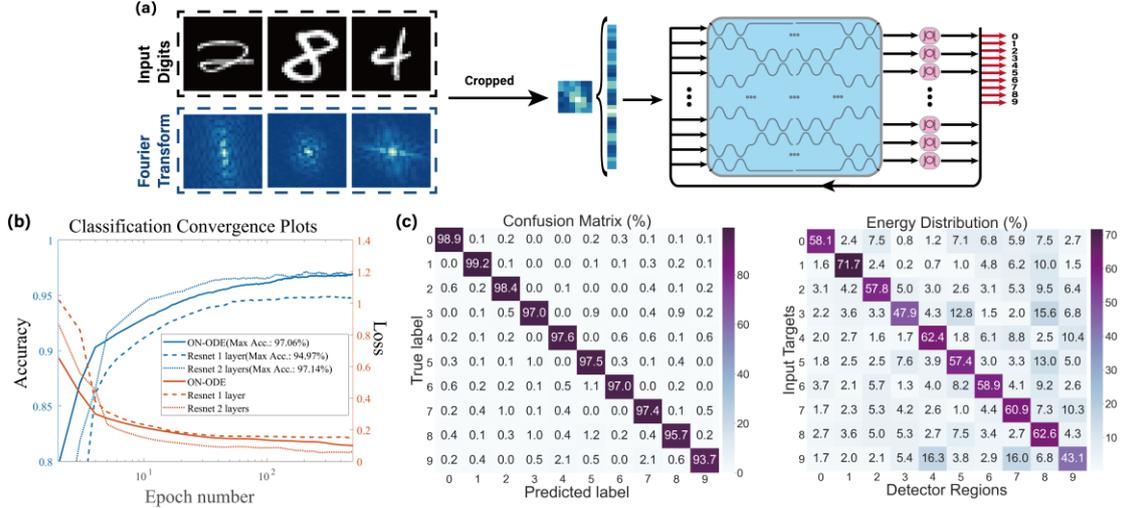

**Fig. 2.** (a) Classifying the MNIST database with ON-ODE configured with optoelectronic nonlinear MZI-based PNNs. The input digits are Fourier-transformed, cropped, and flattened into 36-dimensional vectors to be fed into the ON-ODE configured with a single hidden layer. The network outputs have ten optical waveguides, each corresponding to a digit type. (b) Classification accuracy and loss function with respect to the epoch number during the training of ON-ODE and optical ResNet. (c) Confusion matrix and energy distribution matrix of ON-ODE with a single hidden layer

We perform image classification in the MNIST dataset with1 our ON-ODE. The MNIST dataset contains 70,000 handwritten digits images from 0 to 9; each image has 28 × 28 grayscale pixels. We split the dataset into a training set with 60,000 images and a test set with 10,000 images. To reduce the training time and facilitate the on-chip implementation, we pre-process the input handwritten digits by down-sampling the spatial resolution. We transform the input digits from the spatial domain into the Fourier domain using a 2-dimensional Fourier transform. Several exemplar input digits and their Fourier transformed amplitude images are depicted in Fig. 2(a), where the major information of handwritten digits is located in the central region of the Fourier-domain images. Therefore, we crop the central regions with 6 × 6 pixels and use the low-frequency parts of Fourier images to preserve as much of the spatial information in the simulation. The central regions are flattened into one-dimension signals as the inputs to the on-chip ON-ODE.

For efficiency, in our ON-ODE design, the number of input/output ports exceeds that of the input pixel, which allows taking full advantage of resources on the chip by using almost all the MZIs. As shown in Fig. 2(a), we construct the ON-ODE with 72 input/output ports, where the 2-dimensional handwritten inputs are flattened and fed into the middle 36 ports, and the other 36 ports are set to zero value. Ten output ports are used to represent ten categories of digits types, and the cross-entropy loss function is adopted during the training. The batch size is set to be 512. We train the ON-ODE with a single hidden layer and two hidden layers on the MNIST handwritten digit dataset, which are compared with the optical ResNet with the same numbers of neural network layers and parameters. We set the learning rate to 0.01, and batch size to 100. The Fig. 2(b) shows the training accuracy and the cross entropy loss with respect to the epoch number during the training process.

Empirically, we observe that the optical ResNet with one hidden layer can achieve 94.97% classification accuracy, and adding one more hidden layer can improve the accuracy to 97.14%. Remarkably, the proposed ON-ODE achieve an accuracy of 97.06% with only one hidden layer, which is comparable with the latter. We can figure out that the ON-ODE with one hidden layer can reach a comparable accuracy with respect to optical ResNet with two hidden layers. Suppose we implement the structure of ON-ODE on the chip. In that case, we only need one hidden layer to reach a comparable accuracy of two hidden layers of optical ResNet and significantly reduce the chip area occupancy. Moreover, the ON-ODE is able to take full use of the computing resources compared with ResNet. The confusion matrix and energy distribution matrix of one hidden layer ON-ODE are shown in Fig. 2(c).

In the second task, we use ON-ODE to learn the time series dynamics function. More specifically, we fit an ODE function from points sampled from certain dynamics to restore the trajectory. The structure of ON-ODE is the same as Fig. 2(a), except the input of each port is a complex value encoding the point's 2D coordinates. The input to ON-ODE is the initial value of the trajectory. The initial value is converted into a complex value and is duplicated to be inputted to each port of ON-ODE. We test the complex number of each output port and calculate the average of the real and imaginary parts. These results will be used as the coordinates of the output points to restore the trajectory of the dynamics function.

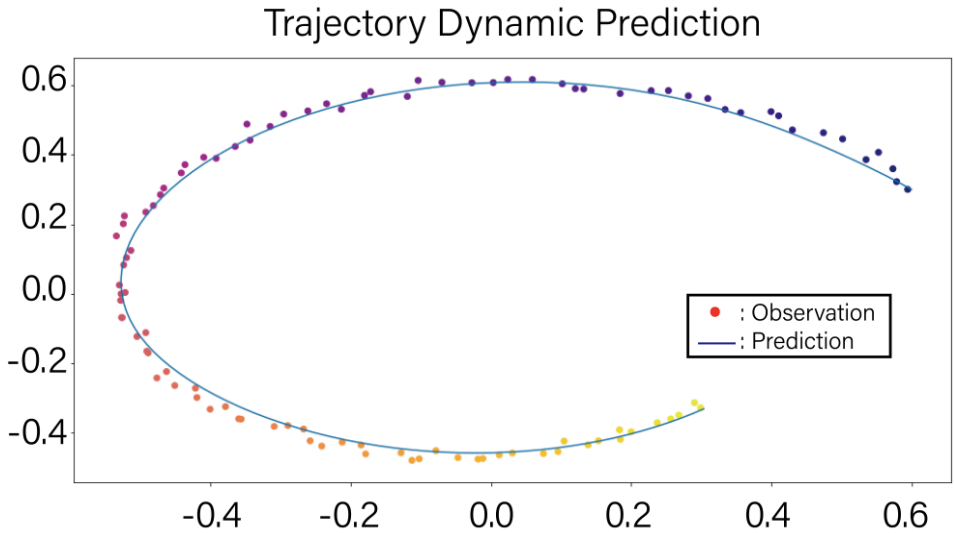

**Fig. 3.** Result of ON-ODE in learning the trajectory dynamics prediction.

An exemplar dynamics function we used for training ON-ODE is:
$$\frac{dx}{dt} = \begin{bmatrix} -0.1 & -1.0 \\ 1.0 & -0.1 \end{bmatrix} x, \tag{4}$$

where $x$ represents the 2D coordinates. The exemplar dynamics function (Equation 4) can be seen as a simplified form of Harmonic oscillator equation, that plays an important role in classical mechanics, RLC electronic circuits, and so on [29]. We first set the initial coordinate $x$ of trajectory to (0.6, 0.3). Then we compute a series of trajectory points based on the equation 4. We train the ON-ODE with 200 points sampled from the trajectory and restore the trajectory after the training process. The loss function used here is the mean squared error (MSE). In the simulation, we train ON-ODE with 2 layers and 9 input/output ports. The training result is depicted in Fig. 3. We can see the proposed ON-ODE could restore the true dynamics function successfully with the MSE of $1.5 \times 10^{-4}$.

We further use the DPU-based all-optical PNNs for constructing the ON-ODE and applied it for object classification. We train phase-only diffractive layers for the image classification task. In the training process, input images are also transformed into the Fourier domain and cropped to 4 × 4 = 16 pixel size in the central regions of images, which is the low-frequency part and retains more information.

These 16 pixels of images are flattened, and we pad these images to 50 pixels with zeros. We upsample these images with the nearest neighbor method to 400 pixels that fit for the DPU. In the article, we classify the images by training a 4-layer DPU, which has 4 × 400 = 1600 trainable parameters and maintain relatively the same numbers of parameters as the MZI-based PNNs.

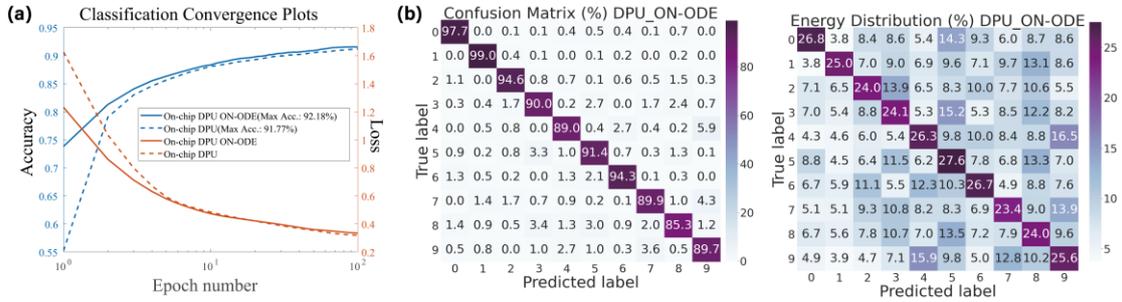

**Fig. 4**. (a) Training accuracy and loss function curve with respect to the epoch number during the training process of DPU-based all-optical PNNs with and without ON-ODE. (b) Confusion matrix and energy distribution matrix of ON-ODE with DPU-based PNNs.

We train the ON-ODE with DPU-based PNN and pure DPU-based PNN on the MNIST handwritten dataset. The training accuracy and cross-entropy loss function curve are shown in Fig. 4(a). From the picture, we conclude that ON-ODE with DPU gains a noticable accuracy improvement over DPU (92.18% vs. 91.77%). It also suggests that ON-ODE can be successfully integrated to both MZIs and DPU and produces satisfying performance. The confusion matrix and energy distribution matrix of ON-ODE networks with DPU are shown in Fig. 4(b). Moreover, we believe that the advantage of computation speed in optical structure could remove the tradeoff between computation speed and accuracy.

We have introduced a novel optical neural network called optical neural ordinary differential equation (ON-ODE). The ON-ODE can be integrated into the DPU-based and MZI-based PNNs with optical integrator and optical feedback loop. We have shown the utility of ON-ODE in image classification and trajectory dynamics recovery. Remarkably, our ON-ODE with one hidden layer performs on par with the two-layer optical ResNet in the first task, which indicates the effectiveness of our design in reducing chip area occupancy. Overall, compared with the electric neural networks, our ON-ODE enjoys higher computation speed and power efficiency as the other PNNs. Furthermore, our ON-ODE requires fewer spaces and resources on the chip than the previous PNN designs, rendering its potential in the future on-chip implementation.

**Funding.** This work is supported by the National Natural Science Foundation of China (No. 62088102), the National Key Research and Development Program of China (No. 2020AA0105500 and No.2552021ZD0109902), and the Tsinghua University Initiative Scientific Research Program.

**Disclosures.** The authors declare no conflicts of interest.

**Data availability**. Data underlying the results presented in this paper are not publicly available at this time but may be obtained from the authors upon reasonable request